\documentclass[11pt,letterpaper]{article}

\usepackage[margin=1in]{geometry}
\usepackage{times}
\usepackage{amsmath}
\usepackage{amsfonts}
\usepackage{amssymb}
\usepackage{graphicx}
\usepackage{natbib}
\usepackage{url}
\usepackage{hyperref}
\usepackage{booktabs}

\providecommand{\keywords}[1]{\noindent\textbf{Keywords:} #1\par}

\title{\LARGE \bf Evaluating Causal Explanation in Medical Reports with LLM-Based and Human-Aligned Metrics}

\author{
    Yousang Cho\\
    Department of Medical Engineering, Konyang University\\
    \texttt{24856504@konyang.ac.kr}
    \and
    Key-Sun Choi\\
    Department of Artificial Intelligence, Konyang University\\
    \texttt{kschoi@\{konyang, kaist\}.ac.kr}, 
}

\date{}

\begin{document}
\maketitle

\begin{center}
\textbf{LLM4Eval Workshop, SIGIR 2025}\\
\textit{Padova, Italy, July 17, 2025}
\end{center}

\vspace{0.2cm}

\begin{abstract}
This study investigates how accurately different evaluation metrics capture the quality of causal explanations in automatically generated diagnostic reports. We compare six metrics—BERTScore, Cosine Similarity, BioSentVec, GPT-White, GPT-Black, and expert qualitative assessment—across two input types: observation-based and multiple-choice-based report generation. Two weighting strategies are applied: one reflecting task-specific priorities, and the other assigning equal weights to all metrics. Our results show that GPT-Black demonstrates the strongest discriminative power in identifying logically coherent and clinically valid causal narratives. GPT-White also aligns well with expert evaluations, while similarity-based metrics diverge from clinical reasoning quality. These findings emphasize the impact of metric selection and weighting on evaluation outcomes, supporting the use of LLM-based evaluation for tasks requiring interpretability and causal reasoning.
\end{abstract}


\section{Introduction}

In the medical field, interest in the application of artificial intelligence (AI) continues to grow, particularly in areas such as automated generation of diagnostic reports. The development of large language models (LLMs) has enhanced the capacity for natural language generation and has begun to influence documentation processes that involve structured data or image-based clinical inputs. In fields such as radiology, internal medicine, and pathology, automated report generation is increasingly regarded as a viable tool to reduce clinician workload and promote standardization. While early applications focused primarily on summarizing observations, there is a growing need to generate reports that include causal explanations as well \cite{hartung2020create}.

A diagnostic report must consist of clinically interpretable information, going beyond listing findings to logically explain how such observations contribute to a diagnostic conclusion. Medical professionals benefit not simply from a sentence listing numerical or descriptive findings, but from a report that contextualizes those findings within diagnostic reasoning. For example, the phrase ``mild bilateral lower lung opacity'' is commonly used, but without linking this observation to a condition such as pneumonia, the report may have limited value. The connection between observations and clinical interpretation is essential for the report to be usable in real clinical practice.

Many existing report generation systems emphasize grammatical fluency and structural consistency but face challenges when it comes to reconstructing causal reasoning or capturing diagnostic logic \cite{waite2017interpretive}. Likewise, the automatic evaluation of generated reports often relies on metrics that were not designed to assess medical reasoning. BLEU, ROUGE, BERTScore, and cosine similarity are commonly used, focusing on lexical overlap or embedding-based semantic similarity. These metrics reflect some aspects of informativeness and phrasing but fail to capture diagnostic relevance or causal coherence, particularly in domain-specific clinical settings.

In clinical documents, a term's meaning can vary significantly depending on context. For example, the word ``consolidation'' may have different implications depending on its location and accompanying features. This makes it difficult to evaluate the clinical accuracy or interpretability of generated text using generic similarity measures alone. Medical reports are interpretive in nature, and thus require evaluators that can assess logical connections and the progression of reasoning.

In response to these challenges, recent approaches have proposed using language models themselves as evaluators. GPT-based models, for instance, can be used not only to generate but also to assess reports by identifying inconsistencies, omissions, or vague explanations. By inputting both the reference and generated reports, models can be prompted to evaluate elements such as coherence, explanatory clarity, and causal appropriateness. This opens possibilities for evaluating meaning beyond surface-level similarity. However, these methods remain in the early stages of validation, and their alignment with expert clinical judgment has not been fully established.

Even as automated report generation technologies advance, discussion continues around how best to evaluate the quality of such output \cite{brady2017error}. There is increasing recognition that evaluation must extend beyond numerical scoring to encompass clinical relevance, the structure of explanations, and the plausibility of diagnostic reasoning. A generated report that correctly names a diagnosis but lacks a supporting explanation or contains logical gaps may not be clinically acceptable. This highlights the need for multi-faceted evaluation frameworks that account for the structure, reasoning, and completeness of medical narratives.

This study considers not only similarity-based metrics but also domain-specific embedding approaches and LLM-based contextual evaluations to assess the quality of reports that include causal explanations \cite{Camburu2018esnli}. We examine how each metric responds to specific aspects of generated reports and compare the consistency of results across different evaluation strategies, with the aim of reducing ambiguity in report assessment and enhancing interpretability of evaluation outputs.

\section{Evaluation Metrics}

To assess the quality of medical reports generated by language models, this study adopted six evaluation metrics. These metrics fall into three general categories: automatic similarity-based evaluations, language model-based evaluations, and expert qualitative evaluation.

\paragraph{BERTScore.}
BERTScore computes semantic similarity between generated and reference sentences using token embeddings from a pre-trained BERT model. It evaluates token-to-token cosine similarity and combines precision, recall, and F1-score into a single metric.

\paragraph{Cosine Similarity.}
Cosine Similarity converts entire sentences into vectors and measures the angle between them to produce a scalar similarity value. While effective in capturing surface-level similarity, it has limitations in reflecting diagnostic reasoning or logical causal structure.

\paragraph{BioSentVec.}
This is a domain-specific sentence embedding model trained on large biomedical corpora such as PubMed and MIMIC \cite{chen2019biosentvec} \cite{Johnson2019mimic}. It shows improved sensitivity to medical terminology and context but remains focused on surface-level semantic alignment.

\paragraph{GPT-based Evaluations.}
GPT-White evaluates surface-level features such as fluency, grammar, informativeness, and clarity. GPT-Black, by contrast, focuses on logical structure, coherence of causal reasoning, and key diagnostic content inclusion. Both provide numerical scores and are prompted with structured queries.

\paragraph{Expert Qualitative Assessment.}
Expert evaluators with prior experience in clinical report evaluation reviewed generated reports for completeness, clarity, and logical coherence. These assessments serve as qualitative benchmarks to assess the reliability of automatic metrics.

\section{Experimental Setup}
This study was structured to compare and analyze various evaluation metrics using a set of pre-collected medical reports. The diagnostic reports evaluated in this study were not generated by the authors themselves. Instead, they were produced by multiple external teams participating in the shared task, based on predefined input formats, and subsequently used as the primary evaluation targets.

\subsection{Report Composition}
The input information was categorized into two distinct types. The first type involved descriptive observations derived from chest radiograph interpretations, requiring models to generate diagnostic reports that included medically meaningful causal explanations. The second type consisted of multiple-choice responses provided by physicians, where the task was to generate an explanatory report that clinically justified the selected answers. In both types, the emphasis was not merely on restating information, but on constructing explanations that reflect medical reasoning and diagnostic coherence.

The reports evaluated in this study were generated by independent teams using various systems. Some models were fine-tuned versions of pre-trained language models, while others were developed based on prompt engineering strategies. Each team received the same input data and submitted the generated reports along with details on the conditions and models used. The authors of this paper were not involved in the report generation process and focused solely on the evaluation phase.

While a structured flow from observation to interpretation and causal explanation was recommended, the actual format and narrative style of the reports varied depending on the model configuration and team-specific strategies.

\subsection{Evaluation Protocol}
This section describes how the six evaluation metrics were applied in the actual report assessment process. Each metric differs in its input requirements, scoring methodology, and output structure. Notably, GPT-based evaluations were tailored based on the type of input used.

Similarity-based metrics (BERTScore, Cosine Similarity, BioSentVec) were applied by matching each generated report to a reference report authored by domain experts. Sentence-level or document-level semantic similarity was calculated automatically and then normalized to enable fair comparison across reports.


\paragraph{GPT-White.}
GPT-White follows a structured, rubric-based evaluation system grounded in publicly available guidelines. It defines explicit criteria and sub-scores across both general and input-type-specific components. Unlike fluency-oriented models, GPT-White emphasizes the precision, completeness, and diagnostic centrality of the content presented in each report. All scores are returned as raw numeric values without explanation, enabling scalable and standardized comparison across reports.

The evaluation rubric is composed of the following components:

\begin{enumerate}
  \item \textbf{Common Evaluation (30 points)}
  \begin{enumerate}
    \item \textit{Contextual Similarity (15 points):} Assesses how well the generated report semantically aligns with the diagnostic and observational content in the reference.
    \item \textit{Diagnosis-Centric Focus (5 points):} Measures the proportion of diagnostic content in the report.
    \item \textit{Adherence to Diagnostic Basis (10 points):} Evaluates how accurately the diagnostic basis from the reference is reflected in the generated report.
  \end{enumerate}
  
  \item \textbf{Input-Type-Specific Evaluation (70 points)}
  \begin{enumerate}
    \item \textit{Observation-Based Report Evaluation}
    \begin{itemize}
      \item Observation Accuracy (20 points)
      \item Clinical Interpretation Appropriateness (20 points)
      \item Clarity and Consistency of Causal Explanation (30 points)
    \end{itemize}
    
    \item \textit{Multiple-Choice-Based Report Evaluation}
    \begin{itemize}
      \item QA4 Diagnostic Mapping Accuracy (20 points)
      \item Causal Reasoning and Clinical Justification (50 points)
    \end{itemize}
  \end{enumerate}
\end{enumerate}
Overall, GPT-White evaluates a report based on diagnostic focus, clarity of medical content, and explanatory completeness. Final scores are automatically calculated as a sum of individual numeric scores.


\paragraph{GPT-Black.}
GPT-Black uses a rule-based bonus and penalty system \cite{chiang2023can} to evaluate the causal integrity and logical coherence of diagnostic reports. In contrast to similarity-based metrics, it assesses whether the report accurately restores causal relationships, avoids clinical errors, and includes key diagnostic reasoning. Evaluations are performed by prompting the model with both the reference and generated reports.

Each report is scored on a scale from 0 to 1 using structured evaluation criteria. The system applies additive bonuses and penalties based on specific rubric rules. Importantly, GPT-Black responses must adhere to a strict numeric-only output format for consistency and reproducibility.

Evaluation prompts are tailored by input type:

\begin{enumerate}
  \item \textbf{Observation-Based Report Evaluation}
  \begin{enumerate}
    \item Contextual Similarity
    \item Logical Consistency of Causal Explanations
    \item Inclusion and Accuracy of Relevant Diagnoses
  \end{enumerate}

  \item \textbf{Multiple-Choice-Based Report Evaluation}
  \begin{enumerate}
    \item QA4 Diagnostic Accuracy
    \item Validity and Completeness of Causal Reasoning
    \item Internal Consistency of Clinical Interpretation
  \end{enumerate}
\end{enumerate}
Each of these aspects contributes to a numerical score, with typical bonus values of +0.2 and penalties of –0.2 or –0.1 depending on the severity of the error. This approach allows GPT-Black to act as a fine-grained evaluator of clinical logic and explanatory completeness.

Expert qualitative evaluation was conducted independently by researchers with prior experience in clinical report assessment. Evaluators did not have access to model identities and based their reviews solely on the input and generated report. The focus was on coherence, clarity, and diagnostic plausibility. Rather than fixed numerical scoring, expert reviews served as qualitative benchmarks for interpreting the reliability of automatic metrics.

\section{Results}

This section presents a detailed quantitative analysis of the performance of multiple models across two distinct input types: Observation-Based Report Evaluation and Multiple-Choice-Based Report Evaluation. Six evaluation metrics—BERTScore, Cosine Similarity, BioSentVec, GPT-White, GPT-Black, and expert qualitative assessment—were employed to capture diverse dimensions of report quality, such as surface-level similarity, contextual alignment, clinical plausibility, and explanatory completeness.

Evaluation scores were computed using two weighting schemes. The first placed higher importance on causal explanation and clinical relevance: 25\% each for GPT-White and GPT-Black, 20\% for BioSentVec and expert evaluations, and 5\% for BERTScore and Cosine Similarity. The second applied equal weights (1/6) to all six metrics, enabling neutral comparison across systems.

The models analyzed were anonymized and referred to as Model A through Model E. These systems were sourced from prior research efforts ~\cite{choi-hiddenrad-taskoverview2024} in automated report generation and submitted outputs based on identical inputs, allowing for standardized evaluation.

\subsection{Observation-Based Report Evaluation}

This task assessed how well models could reconstruct diagnostic reasoning based on descriptive findings in radiology reports. Table~\ref{tab:observation} displays evaluation results.

\begin{table}[h]
  \centering
  \caption{Evaluation scores across six metrics and two weighting schemes for Observation-Based Report Evaluation}
  \label{tab:observation}
  {\footnotesize
  \begin{tabular}{lccccccc}
    \toprule
    Model & BERT & CosSim & BioSent & GPT-W & GPT-B & Expert & Wtd / Eq \\
    \midrule
    A & 0.281 & 0.570 & 0.785 & 0.696 & 0.715 & 0.689 & 0.690 / 0.623 \\
    B & 0.236 & 0.522 & 0.770 & 0.691 & 0.713 & 0.694 & 0.680 / 0.604 \\
    C & 0.256 & 0.541 & 0.766 & 0.680 & 0.700 & 0.690 & 0.678 / 0.606 \\
    D & 0.259 & 0.538 & 0.767 & 0.683 & 0.696 & 0.682 & 0.677 / 0.604 \\
    E & 0.179 & 0.571 & 0.765 & 0.633 & 0.689 & 0.694 & 0.660 / 0.589 \\
    \bottomrule
  \end{tabular}
  }
\end{table}

Model A achieved the highest performance in GPT-White (0.696), GPT-Black (0.715), BioSentVec (0.785), and expert review (0.689), resulting in top scores under both weighting strategies (0.690 and 0.623). These outcomes indicate that Model A provides coherent and clinically accurate causal narratives.

Models B and C showed closely aligned profiles, but their rankings varied with the weighting strategy. Under task-prioritized weights, Model B slightly outperformed Model C (0.680 vs. 0.678), while the order reversed under equal weighting due to stronger BERTScore and Cosine Similarity from Model C.

Model D’s profile closely matched that of Model C, with solid GPT-based and BioSentVec scores. Its marginally lower performance in BERTScore and expert evaluation led to slightly reduced total scores.

Model E demonstrated relatively stable performance in GPT and expert evaluations, but weak BERTScore (0.179) and GPT-White (0.633) reduced its overall rankings, pointing to issues in sentence-level structure and diagnostic clarity.

Overall, the results show that GPT-based metrics and expert evaluations are crucial for assessing causal integrity and clinical coherence, while surface-level metrics such as BERTScore and Cosine Similarity have limited discriminative utility in this context.

\subsection{Multiple-Choice-Based Report Evaluation}

This task tested the models’ ability to produce diagnostically valid and causally coherent reports based on structured question-answer (QA) inputs. Table~\ref{tab:mc} presents the results.

\begin{table}[h]
  \centering
  \caption{Evaluation scores across six metrics and two weighting schemes for Multiple-Choice-Based Report Evaluation}
  \label{tab:mc}
  {\footnotesize
  \begin{tabular}{lccccccc}
    \toprule
    Model & BERT & CosSim & BioSent & GPT-W & GPT-B & Expert & Wtd / Eq \\
    \midrule
    A & 0.099 & 0.669 & 0.827 & 0.827 & 0.859 & 0.816 & 0.790 / 0.683 \\
    B & 0.123 & 0.590 & 0.762 & 0.798 & 0.788 & 0.780 & 0.740 / 0.640 \\
    C & 0.224 & 0.634 & 0.778 & 0.740 & 0.723 & 0.783 & 0.720 / 0.647 \\
    \bottomrule
  \end{tabular}
  }
\end{table}

Model A again demonstrated exceptional performance, securing the highest scores in GPT-White (0.827), GPT-Black (0.859), and expert review (0.816), which translated into leading rankings under both weighting schemes (0.790 and 0.683).

Model B was strong in GPT-based metrics and expert evaluation, with a minor decline in BERTScore (0.123) and Cosine Similarity (0.590). It placed second in prioritized weighting (0.740), but was overtaken by Model C under equal weighting (0.640 vs. 0.647).

Model C scored highest in BERTScore (0.224) and Cosine Similarity (0.634) while lagging slightly in GPT evaluations. These surface metrics gained greater influence under equal weighting, elevating its overall score.

GPT-Black showed the broadest score range (0.136 between top and bottom models), underlining its effectiveness in assessing the depth and structure of causal explanation. GPT-White and expert scores showed parallel trends, reinforcing their validity.

BERTScore, although high for Model C, diverged from GPT and expert evaluations, indicating it may emphasize lexical overlap over diagnostic quality. Cosine Similarity exhibited narrow variance and minimal impact.

\subsection{Metric Comparison and Core Indicator Analysis}

Metric-specific comparisons across both tasks confirmed that GPT-Black is the most discriminative metric. In the observation task, its top–bottom score gap was 0.026; in the multiple-choice task, the gap widened to 0.136.

GPT-White also demonstrated high discriminative power with consistent alignment to expert evaluation, highlighting its utility in structured rubric-based scoring. Expert qualitative assessments mirrored GPT-based rankings in most cases, validating the automation potential.

Conversely, BERTScore and Cosine Similarity often failed to reflect clinical reasoning. For example, Model C’s highest BERTScore (0.224) in the multiple-choice task did not translate into high GPT-Black (0.723) or expert (0.783) scores, underscoring the disconnect between surface similarity and causal soundness.

Overall, GPT-based metrics and expert review emerged as key indicators of clinical quality and explanatory completeness. These should form the backbone of evaluation frameworks for tasks involving diagnostic interpretation. BERTScore and Cosine Similarity may be best used as auxiliary indicators, while BioSentVec offered limited discrimination due to its tight score distribution.

\section{Discussion}

The results of this study highlight several key insights into the evaluation of diagnostic reports that include causal explanations. Specifically, the findings emphasize the varying effectiveness of different evaluation metrics, the influence of metric weighting schemes on final model rankings, and the implications of these results for future evaluation framework design.

\paragraph{Importance of Causal Metrics.}
GPT-Black consistently demonstrated the strongest discriminative power across both input types. Its evaluation of logical coherence, diagnostic completeness, and causal alignment proved vital for identifying reports with sound clinical reasoning. The wide scoring range and sensitivity to logical errors make GPT-Black particularly effective in distinguishing high- and low-quality causal explanations.

GPT-White also contributed significantly to reliable ranking, capturing contextual and structural aspects of explanatory narratives. Its rubric-based scoring system allowed for transparent and replicable evaluations. Notably, its high correlation with expert qualitative scores reinforces its utility as a core component in automated evaluation.

\paragraph{Limitations of Surface-Level Metrics.}
Metrics such as BERTScore and Cosine Similarity, while useful for assessing phrasing and superficial semantic similarity, often failed to reflect the depth or validity of diagnostic reasoning. Their limited score variance and divergence from expert judgment in several cases highlight the risk of over-reliance on these indicators. These findings affirm the necessity of incorporating clinically-informed metrics when evaluating explanation quality.

\paragraph{Effect of Weighting Schemes.}
The analysis also revealed how metric weighting impacts final rankings. Task-specific weighting schemes—prioritizing causal relevance and clinical interpretability—favored models with stronger reasoning capabilities. In contrast, equal weighting schemes increased the influence of surface metrics, leading to rank reversals in cases where causal coherence was not fully captured by similarity measures. These observations underscore the importance of aligning metric weights with evaluative goals.

\paragraph{Expert Evaluation and Future Directions.}
While expert evaluations provided an essential qualitative benchmark, the current study relied on a limited number of reviewers and lacked inter-rater agreement analysis. Future research should implement multi-reviewer protocols, detailed rubric validation, and strategies for managing inter-reviewer variance. Furthermore, incorporating domain-specific knowledge graphs or ontology-driven tools (e.g., SNOMED CT) into evaluation could help bridge the gap between clinical logic and language-based models.

Lastly, evaluation frameworks should evolve to handle more complex report scenarios—such as temporally sequenced explanations, multi-condition reasoning, and differential diagnoses. Hybrid approaches that integrate structured expert feedback with LLM-based assessments offer a promising path forward.

\section{Conclusion}

This study examined the evaluation of causal explanations in diagnostic reports generated by language models using six distinct metrics across two task types. The analysis demonstrated that GPT-based evaluation metrics—particularly GPT-Black and GPT-White—alongside expert qualitative assessment, were the most effective at capturing the logical structure, clinical relevance, and explanatory completeness of generated reports.

GPT-Black emerged as a powerful metric due to its focus on causal coherence and reasoning accuracy. GPT-White complemented this by offering structured, rubric-based assessment with high correlation to expert judgment. These metrics should be prioritized in any evaluation framework where interpretability and reasoning transparency are critical.

In contrast, surface-level metrics such as BERTScore and Cosine Similarity were found to have limited alignment with clinical validity. While useful for capturing lexical and semantic overlap, they lacked sensitivity to diagnostic logic and contextual accuracy. BioSentVec provided domain-specific improvements but exhibited narrow score variance across models, limiting its discriminative potential.

The sensitivity of model rankings to different weighting strategies further reinforced the importance of aligning metric weights with task objectives. Evaluation designs should reflect the intended use of generated reports—whether for clinical support, training, or research—and weigh metrics accordingly.

Future directions include enhancing the robustness of expert evaluations, introducing multi-rater reliability checks, and exploring hybrid evaluation methods that integrate ontology-based tools or knowledge-augmented models. These approaches could address complex report structures and improve the interpretability of AI-generated content in medical applications.

Ultimately, this work contributes a foundation for developing more nuanced and clinically meaningful evaluation methodologies for causal explanation generation in medical AI systems.

\bibliographystyle{plain}
\bibliography{sample-base}

\begin{thebibliography}{1}

\bibitem{brady2017error}
Adrian~P Brady.
\newblock Error and discrepancy in radiology: inevitable or avoidable?
\newblock {\em Insights into imaging}, 8:171--182, 2017.

\bibitem{Camburu2018esnli}
Oana-Maria Camburu, Tim Rockt{\"a}schel, Thomas Lukasiewicz, and Phil Blunsom.
\newblock e-snli: Natural language inference with natural language explanations.
\newblock In {\em Neural Information Processing Systems}, 2018.

\bibitem{chen2019biosentvec}
Qingyu Chen, Yifan Peng, and Zhiyong Lu.
\newblock Biosentvec: creating sentence embeddings for biomedical texts.
\newblock {\em Proceedings of the IEEE International Conference on Healthcare Informatics (ICHI)}, pages 1--5, 2019.

\bibitem{chiang2023can}
Cheng-Han Chiang and Hung-yi Lee.
\newblock Can large language models be an alternative to human evaluations?
\newblock {\em arXiv preprint arXiv:2305.01937}, 2023.

\bibitem{hartung2020create}
Michael~P Hartung, Ian~C Bickle, Frank Gaillard, and Jeffrey~P Kanne.
\newblock How to create a great radiology report.
\newblock {\em Radiographics}, 40(6):1658--1670, 2020.

\bibitem{Johnson2019mimic}
Alistair E.~W. Johnson, Tom~J. Pollard, Seth~J. Berkowitz, Nathaniel~R. Greenbaum, Matthew~P. Lungren, Chih ying Deng, Roger~G. Mark, and Steven Horng.
\newblock Mimic-cxr, a de-identified publicly available database of chest radiographs with free-text reports.
\newblock {\em Scientific Data}, 6, 2019.

\bibitem{choi-hiddenrad-taskoverview2024}
{Key-Sun Choi and Yousang Cho and Hidden-Rad Organizing Committee}.
\newblock Overview of the ntcir-18 hidden-rad task: Hidden causality inclusion in radiology report generation.
\newblock In {\em Proceedings of the 18th NTCIR Conference on Evaluation of Information Access Technologies}, jun 2025.
\newblock forthcoming.

\bibitem{waite2017interpretive}
Stephen Waite, Jinel Scott, Brian Gale, Travis Fuchs, Srinivas Kolla, and Deborah Reede.
\newblock Interpretive error in radiology.
\newblock {\em American Journal of Roentgenology}, 208(4):739--749, 2017.

\end{thebibliography}


\end{document}